\documentclass{article}
\pdfoutput=1
\PassOptionsToPackage{numbers, compress}{natbib}

\usepackage[preprint]{neurips_2023}

\usepackage[utf8]{inputenc} 
\usepackage[T1]{fontenc}    
\usepackage{hyperref}       
\usepackage{url}            
\usepackage{booktabs}       
\usepackage{amsfonts}       
\usepackage{nicefrac}       
\usepackage{microtype}      
\usepackage{xcolor}
\usepackage{algorithm}
\usepackage{algpseudocode}
\usepackage{lipsum}

\usepackage{times}
\usepackage{latexsym}
\usepackage{inconsolata}
\usepackage{amsmath}
\usepackage{mathrsfs}
\usepackage{graphics}
\usepackage[dvipdfm]{graphicx}
\usepackage{xr}
\usepackage{multirow}
\usepackage{tabularx}
\usepackage{balance}
\usepackage{caption}
\usepackage{subcaption}
\usepackage{enumitem}
\usepackage{pgfplots}
\usepackage{tikz}
\usepackage{xspace}
\usepackage{makecell}
\usepackage{amssymb}
\usepackage{amsthm}
\usepackage{graphicx}
\usepackage{calc}
\usepackage{cleveref}

\newtheorem{definition}{Definition}
\newtheorem{theorem}{Theorem}
\newtheorem{lemma}{Lemma}
\newcommand*\samethanks[1][\value{footnote}]{\footnotemark[#1]}

\title{Approximating Human-Like Few-shot Learning \\
with GPT-based Compression}

\author{%
  Cynthia Huang\thanks{equal contribution} \quad Yuqing Xie\samethanks \quad Zhiying Jiang\samethanks \quad Jimmy Lin \quad Ming Li   \\
  Cheriton School of Computer Science, University of Waterloo \\
  \texttt{\{hhuang, yuqing.xie, zhiying.jiang, jimmylin, mli\}@uwaterloo.ca}
}

\begin{document}

\maketitle

\begin{abstract}

In this work, we conceptualize the learning process as information compression. 
We seek to equip generative pre-trained models with human-like learning capabilities that enable data compression during inference. 
We present a novel approach that utilizes the Generative Pre-trained Transformer (GPT) to approximate Kolmogorov complexity, with the aim of estimating the optimal Information Distance for few-shot learning. 
We first propose using GPT as a prior for lossless text compression, achieving a noteworthy compression ratio. 
Experiment with LLAMA2-7B backbone achieves a compression ratio of 15.5 on enwik9.
We justify the pre-training objective of GPT models by demonstrating its equivalence to the compression length, and, consequently, its ability to approximate the information distance for texts. 
Leveraging the approximated information distance, our method allows the direct application of GPT models in quantitative text similarity measurements. 
Experiment results show that our method overall achieves superior performance compared to embedding and prompt baselines on challenging NLP tasks, including semantic similarity, zero and one-shot text classification, and zero-shot text ranking.

\end{abstract}

\section{Introduction}

Large labeled datasets are often scarce in the real world where annotation is expensive and time consuming. This has prompted the development of few-shot learning, where the model is learned using only a few annotated samples~\cite{finn2017model}.
One resort to the few-shot scenario is to utilize the pre-trained models like Generative Pre-trained Transformers (GPTs)~\cite{gpt, radford2019language, brown2020language, openai2023gpt4} with in-context learning~\cite{brown2020language, zhao2021calibrate}, fine-tuning~\cite{liu2022few} or the combination~\cite{ben2022pada}. However, in-context learning requires heavy engineering to achieve a high accuracy~\cite{liu2023pre}, and its ability to generalize to different tasks is constrained by the input size and the need for precise formatting. Fine-tuning also has limitations, notably its inability to generalize to rare out-of-distribution datasets with limited labeled samples~\cite{yu2021empirical,nogueira2020document}.

Contrary to the data-hungry nature of machine learning, humans demonstrate exceptional ability in zero-shot and few-shot learning, where only a handful of labeled examples are needed to grasp a concept. Inspired by this, \citep{humanlikefewshot} proposed a human-like few-shot learning framework and boils it down to the ability of compressing data at inference time, measured by the universal information distance. Derived from a simple and accepted assumption in thermodynamics \cite{bennett1998information}, the universal information distance emerges as the key component in taking the effective usage of few labeled data. Hence, accurate approximations of the information distance can lead to improved learning, though its incomputability has posed significant challenges.
This information distance, consisting of Kolmogorov complexity~\cite{Li2008-LIAIT-3}, is both data-type-agnostic and distribution-agnostic. 
Additionally, its parameter-free usage enables the metric's applicability across diverse applications.
Inspired by this theory, we propose a novel method GPT-AC, that leverages the knowledge GPTs learned during pre-training. Our method tackles tasks traditionally challenging for prompt engineering or fine-tuning, including semantic similarity, zero-shot text classification and ranking.

Kolmogorov complexity, also known as the length of the shortest computer program for a target output, is often approximated by the compression length. At the same time, entropy coding algorithms attempt to approach the code length lower bound declared by Shannon's source coding theorem: the entropy of the source. By using rich priors, such as pre-trained language models, that more accurately predict the source distribution, we can optimize the compression ratio and more closely approximate Kolmogorov complexity.
Despite the potential of large language model-based compressors, their direct application to downstream tasks is nearly infeasible due to speed constraints. In addition to the inference speed required by the language model itself, the overhead of the compressor is even more substantial. 
Fortunately, the information distance only requires the compressed length instead of actual compression of the text sequence.
We demonstrate an equivalence of the compression length under arithmetic coding to the negative log probability of text tokens when using GPT as the prior. This easy-to-compute compression length enables us to efficiently approximate the universal information distance without the overheads of the actual compression. By approximating normalized information distances \cite{ncd, nid1, li2004similarity} using GPT-based compression, we significantly enhance GPT's ability to quantitatively capture text similarities, which forms the foundation for its application in downstream NLP tasks.

\vspace{-0.1cm}
Our contributions are as follows: 
\begin{enumerate}[itemsep=0pt,topsep=0pt,parsep=0pt,partopsep=0pt, leftmargin=12pt]
\item We propose a novel method that utilizes generative pre-trained language models to approximate information distance for few-shot learning without the need for fine-tuning or prompt engineering.
\item By connecting arithmetic coding's compression length to the cumulative negative log probabilities in GPT models, we efficiently compute and approximate the information distance derived from Kolmogorov complexity.
\item We validate the effectiveness of our method through experiments in semantic textual similarity, text classification and re-ranking under zero-shot and one-shot settings, exhibiting notable improvement over embedding and prompt baselines.
\item We also demonstrate that our lossless text compression method GPT-AC achieves SOTA compression ratio with Llama2-7B backbone, highlighting the potential of pre-trained large language models as powerful priors in compression.
\end{enumerate}

\section{Related Works}

\subsection{Few-shot Learning}

Prior to the emergence of large pre-trained models, the majority of previous works on few-shot learning can be divided into two streams: meta/transfer-learning based methods~\cite{vinyals2016matching, edwards2016towards, snellprototypical, sung2018learning} and data augmentation based methods~\cite{kwitt2016one, pfister2014domain, douze2018low, gao2018low}. However, the former relies on constraining the hypothesis space by prior knowledge from other tasks or support datasets while the latter depends on the heuristics of the datasets, often accompanied with the smoothness assumption~\cite{van2020survey} (i.e., close data points in the input space share the same label).
Pre-trained models, on the other hand, have incorporated prior knowledge during the pre-training stage and are proved to be excellent few-shot learners~\cite{brown2020language}. However, pre-trained models suffer from (1) high computational cost and (2) unsatisfactory performance in out-of-distributed datasets~\cite{yu2021empirical}. The problem of computational cost is especially prominent for large language models like GPT-3 where it is infeasible to fine-tune locally. We utilize pre-trained language model for one-shot and zero-shot classification tasks with no fine-tuning required.

\subsection{Kolmogorov Complexity and Compression Distance}

Information distance was first proposed by~\cite{bennett1998information} as a universal similarity metric between two objects. It was proved to be universal in a way that it is optimal and can discover every possible metric~\cite{bennett1998information}. Due to the problem of incomputability, ~\cite{chen2004shared, li2004similarity, ncd} have derived computable version of information distances for tasks like clustering and plagiarism detection, shedding light on the possibility of using real-world compressors to approximate Kolmogorov complexity. These prior works utilize traditional compressors, indicating that the performance gain on downstream tasks mainly comes from the compressor-based distance metric.

Recently, ~\cite{fewshotnips} propose non-parametric learning by compression with latent variables (NPC-LV) where neural networks are incorporated into information distance. 
They demonstrate that trained generative models like variational autoencoders can be used directly with zero parameters for downstream few-shot image classification. 
Also, ~\cite{jiang-etal-2023-low} employ this method in text classification using GZIP, achieving competitive results compared to several widely-used deep learning approaches.
However, it remains open in how to incorporate pre-trained language models into this framework, which we aim to address in this paper.
A recent study ~\cite{wu2023selfadaptive} explores the use of compression length for  
in-context example selection. They rely on a large candidate set and applied the model under generation setting. In contrast, we focus on adapting generative models to zero/one-shot learning for text similarity tasks.

\subsection{Neural Compression}
Our GPT-based compressor falls under the category of neural compression where neural networks are used for data compression.
Shannon's source coding theorem~\cite{shannon1948mathematical} establishes the lower bound of the lossless compression on random variables with probability distribution. With near-optimal coding schemes, the bottleneck is the entropy approximation.
Fortunately, deep generative models with explicit density estimation serve as the entropy model that can learn adaptively. 
~\cite{bitsbackencoding} propose Bits-Back with Asymmetric Numeral Systems (BB-ANS), a lossless compression algorithm based on VAE. 
Bit-Swap~\cite{BitSwap} further improves BB-ANS by incorporating multiple latent variables and hierarchical networks. In addition to autoencoders, Flow~\cite{rezende2015variational,wang2022fast}-based lossless compression~\cite{hoogeboom2019integer} outperform Bit-Swap and achieve the state of the art compression ratio of images.
The development of deep neural networks also benefits lossless text compression. ~\cite{goyal2019deepzip} use recurrent neural networks~\cite{schuster1997bidirectional} combining with arithmetic coding~\cite{Arithemticcoding} to achieve higher compression ratio than GZIP. 
Recent advancements, such as the fast transformer-based general-purpose lossless compressor TRACE ~\cite{trace}, have demonstrated promising results in enhancing compression performance with transformer architecture.
NNCP~\cite{nncp} v3.1, adaptively encodes the source message with Transformers, achieves state-of-the-art performance on the Matt Mahoney's Large Text Compression Benchmark\footnote{\url{http://mattmahoney.net/dc/text.html}}.

\subsection{Pre-trained Models}
Pre-training has been adopted in numerous deep learning models with the rise of transformer~\cite{vaswani2017attention} due to its ability of learning task-agnostic representation. In NLP, encoder-only transformers like BERT~\cite{devlin2019bert} has achieved impressive performance on GLUE benchmark~\cite{wang2018glue} including tasks like natural language inference and sentiment analysis with only MLP and fine-tuning. Decoder-only transformers like GPT~\cite{radford2019language,brown2020language} can treat downstream discriminative tasks in a generative way. However, previous works on few-shot learning using language models are either prompt-based~\cite{petroni2019language,perez2021true,li2021prefix} or fine-tuning-based~\cite{zhao2021calibrate,yamada2020luke,nogueira2020document} while in this work, we propose a new way to leverage pre-trained language models for few-shot learning without fine-tuning or in-context learning.

\section{Method}

\subsection{Human-Like Few-Shot Learning and Universal Information Distance}

We consider the following generalized human-like few-shot learning setting: assume a universe of objects $\Omega$ comprising various concept classes. Given an unlabelled subset $U \subset \Omega$, a hypothesis $\phi$ is formulated. For any concept class $c$, we have information $D_c$ representing either a description or representations derived from a few examples. 
The goal is to determine the concept class for any new instance $x \in \Omega$ under a computable metric $\mathcal{M}$:
\begin{equation}
    argmin_{c \in \mathcal{C}} \mathcal{M}(x, D_c | \phi(U)).
\end{equation}

Here, $\phi$ can be a pre-trained model that learned the distribution from the unlabeled data $U$. For instance, GPT can be seen as the hypothesis $\phi$ learned from a large unlabeled corpus.

This learning scenario differs from traditional machine learning settings as it permits extensive pre-training using unlabeled data but provides very limited labeled data for learning concept classes. Under this framework, the optimal metric $\mathcal{M}$ for few-shot learning algorithms is proven to be the universal information distance, defined by Kolmogorov complexity (details are shown in Appendix), based on the von-Neuman-Landauer Principle. Formally, the universal information distance $\mathcal{M}_{UID}$ between two objects $x$ and $y$ is defined as:
\begin{equation}
    \mathcal{M}_{UID}(x, y) = \max\{K(x|y), K(y|x)\},  \label{eq:origin-kol}
\end{equation}
$K(x|y)$ is the Kolmogorov complexity of $x$ given $y$, or informally, the length of the shortest program that outputs $x$ given input $y$. Since the Kolmogorov complexity is not computable~\cite{complexityincomputable}, it is often approximated via compression in practice. 

\subsection{GPT-AC: Generative Pre-trained Transformer based Arithmetic Coding for Compression}

In this section, we propose GPT-based Arithmetic Coding (GPT-AC) where GPT is integrated into adaptive arithmetic coding, an entropy-based compressor.

\textbf{GPT as the Entropy Model}\\
Consider a text $T = (t_1, t_2, ..., t_n)$  composed of a sequence of tokens. Let $\phi$ represent a GPT model, where $\phi(t_{1:(i-1)}) = P_i(t_i|t_1, t_2, ..., t_{i-1})$ models the probability distribution $P_i$ of the next token $t_i$. The function $\phi(T)$ outputs all next-token probability distributions $(P_2, \cdots, P_{n+1})$. To derive the distribution for $P_1$, an EOS (End Of Sentence) token is added at the start of the text as $t_0$. For each token $t_i$, the associated $P_i$ serves as the entropy model for both encoding and decoding in the compressor.

\vspace{-0.3cm}
\begin{figure}[ht!]
    \centering
    \includegraphics[scale=0.7]{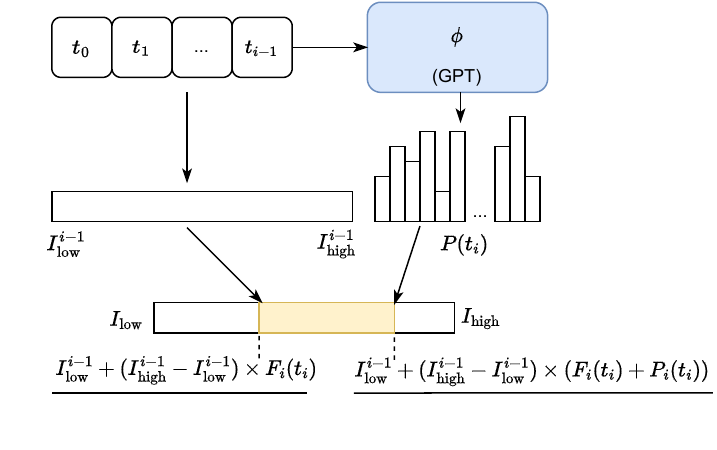}
    \vspace{-0.3cm}\caption{Illustration of GPT-based Arithmetic Encoding}
\end{figure}

\vspace{-0.3cm}
\textbf{GPT-AC Encoding}\\
In the encoding phase, under the scheme of adaptive arithmetic coding, we start with an initial interval $I^0 = [0,1)$. For each token $t_i$ in the sequence, we calculate the cumulative distribution function $F_i(t_i)$ and $P_i(t_i)$ based on $\phi(t_{1:(i-1)})$. Then, the interval $I = [I_\text{low}, I_\text{high})$ is updated according to the range assigned to $t_i$: 
\begin{equation}
\begin{split}
    &I_\text{low}^i = I_\text{low}^{i-1} + (I_\text{high}^{i-1} - I_\text{low}^{i-1}) \times F_i(t_i), \\
    &I_\text{high}^i = I_\text{low}^{i-1} + (I_\text{high}^{i-1} - I_\text{low}^{i-1}) \times (F_i(t_i) + P_i(t_i))
\end{split}
\end{equation}
After updating $I$ for each token in the sequence, we can pick any number $x$ within the final interval to represent the entire text sequence. 

\textbf{GPT-AC Decoding}\\
When decoding the encoded message $x$, the token $t_1$ can be identified by finding the range $[F_1(t_1), F_1(t_1) + P_1(t_1)]$ that includes $x$. The value of $x$ is then updated by normalizing it within the range of $t_1$, using the formula: $x \leftarrow \frac{x - F_1(t_1)}{P_1(t_1)}$. With this updated $x$ and the next-token probability distribution $\phi(t_2)$, we can decode the next token $t_2$. This process is repeated until an EOS token is encountered, indicating the end of the text. The text can be losslessly decoded using $x$ and $\phi$ alone. 

\textbf{Negative Log Probability as Compression Length}\\
During the arithmetic encoding, the length of the interval $I^i$ equals to $I^{i-1} * P_i(t_i)$. From an initial interval of length 1, the entire message's encoding results in a final interval with a length of $\prod^{n}_{i=1}P_i(t_i)$. The number of bits required to represent this final interval, and thus the message $T$, is $\sum_{i=1}^{n} -\log_2 P_i(t_i)$. This reveals a method to approximate the compression length directly without exactly performing the compression. With the triangular forward attention masking in GPT, we can pass the full tokenized text sequence to the model and obtain probability distributions for all tokens. 

\textbf{GPT Pre-training Optimizes for Compression Length} \\
The optimization target during pre-training for auto-regressive models such as GPT is defined as:
\begin{align*}
L(T|p_{model}) &= -\log p_{model}(T) 
            = -\log p_{model}(t_1, t_2, ..., t_n) = \sum_{i=1}^{n} -\log_2 P_i(t_i).
\end{align*}
For entropy coding, $H(T)\triangleq \mathbb{E}(-\log p_{data}(T))$, defining the optimal code length. While $p_{data}$ is often unknown, we thus use the observation $p_{\bar{data}}$ to approximate $p_{data}$:
\begin{align*}
    H(T)    &= \mathbb{E}_{p_{data}}[-\log p_{model}(T)]  
             \simeq \mathbb{E}_{p_{\bar{data}}}[-\log p_{model}(T)]
\end{align*}
According to The Shannon–Fano–Elias coding scheme~\cite{cover1999elements}, we can construct a prefix-free code of length $-\log p_\text{model}(t_1, t_2, ..., t_n)+2$ bits.
Consequently, the pre-training phase of GPT models is essentially learning a compressor that optimizes the coding length.

\textbf{The Rank Coding Variant} \\
In the method outlined above, we primarily employ arithmetic coding for text compression. An alternative variant of lossless coding uses rank instead of probability~\cite{herrera2021}. After the GPT model predicts the distribution of the next token, we can rank all tokens according to their probability. The target token is assigned the corresponding rank index, with higher probability tokens having smaller rank indices. In this variant, we approximate the compression length as $\sum^n_{i=1} \log_2(rank_i)$ where $rank_i$ denotes the rank for token $i$. 

\textbf{Applicability of Our Method} \\
When calculating compression length, the algorithm is most efficient with transformers that use forward-only attention. This approach allows for a single-pass processing of the entire text sequence, thus ensuring efficient computation. However, through the use of a sliding window technique, the method's applicability can be extended to all generative language models covering decoder-only and encoder-decoder architectures.

\subsection{Universal Information Distance Approximation}\label{sec:uida}

Having computed the compression length using the GPT-AC method, we can now utilize it to approximate the universal information distance. Let $x = \{x_1, \cdots, x_n\}$ and $y = \{y_1, \cdots, y_m\}$ denote two tokenized text sequences, where each $x_i \text{ or } y_i$ represents a token in the sequence. We approximate $K(x)$ using the compression length $C(x) = \sum_{i=1}^n -\log_2 {P_i(x_i)}$ where $P_i$ represents the probability distribution for $x_i$ as predicted by the GPT model. 

As in \Cref{eq:origin-kol}, we also need $K(x|y)$, approximated as follows: let $P_i = \phi(y, x_{1:{i-1}})$ denotes the probability distribution for token $x_i$ output by the GPT model, given $y=(y_1, \cdots, y_m)$ and previous tokens in $x$, $K(x|y)$ can be estimated as $C(x|y)=\sum_{i=1}^n -\log_2 {P_i(x_i)}$. A similar approach can be used to estimate $K(y)$ and $K(y|x)$. We denote all compressed-based approximations in $C(\cdot)$.

However, compression lengths vary when the lengths of the input text differ. We need a normalized version to enable comparison across diverse object pairs. There are several normalized measures. 
\cite{li2004similarity} introduced a normalized version, referred to as the Normalized Information Distance (NID):
\begin{equation}
\mathcal{M}_{max}(x, y) = \frac{\max\{K(x|y), K(y|x)\}}{\max\{K(x), K(y)\}}
\end{equation}
To tackle challenges such as partial matching
\footnote{Partial matching means situations where only portions of two objects match each other.}
, \cite{ml2006} proposed the following variants of the universal distances suitable for broader application scenarios:
\begin{equation}
\mathcal{M}_{min}(x, y) = \frac{\min\{K(x|y), K(y|x)\}}{\min\{K(x), K(y)\}}.
\end{equation}
\cite{10.1145/1014052.1014077} proposed the Compression-Based Dissimilarity Measure (CDM) for data mining applications, demonstrating its effectiveness in practice. This is rescaled to fit the range $[0, 1)$:
\begin{equation}
\mathcal{M}_{mean}(x, y) = \frac{C(x|y) + C(y|x)}{C(x)+ C(y)}  = 2 * CDM - 1 \text{,  } CDM =  \frac{C(xy)}{C(x) + C(y)} 
\end{equation}

\subsection{Applications of Universal Information Distance}
We will now explain how the aforementioned distances can be applied to various NLP tasks.
To determine the text similarity score between two texts $x, y$, we first compute their individual compression lengths $C(x), C(y)$. We also calculate the joint and conditional compression lengths $C(xy), C(x|y), C(y|x)$. Using these values, we compute the distance metrics defined in \Cref{sec:uida} as $\mathcal{M}$. We can then apply these distance measures to specific tasks:
\vspace{-0.1cm}
\begin{itemize}[itemsep=0pt,topsep=0pt,parsep=0pt,partopsep=0pt, leftmargin=12pt]
    \item For semantic textual similarity, we treat the two sentences as $x$ and $y$, and use $\mathcal{M}$ as predictions.
    \item For zero-shot text classification, we treat the label descriptions as $x$ and the multiple choice options as $y$.
    For one-shot text classification, we consider the training sample as $x$ and the test sample as $y$. We classify the text sample by comparing $\mathcal{M}$ values for different classes.
    \item For text re-ranking, we treat the documents as $x$ and the queries as $y$, ranking according to $\mathcal{M}$.
\end{itemize}

\section{Experiments}

Our experimental evaluation consists of four key components: lossless text compression and three downstream tasks, namely semantic textual similarity, text classification, and text re-ranking. 
For downstream applications, we mainly conduct experiments with the GPT-2 small (124M),
comparing to GPT-2 embedding or prompt tuning baselines and BERT-base-uncased (110M) models from HuggingFace Hub\footnote{\url{https://huggingface.co/}}.
We take GPT-2 as an example due to its light weights and availability.
However, the proposed method is not limited to GPT models. It can be readily applied to more advanced large language models, such as LLAMA\cite{touvron2023llama}, provided that the output probabilities are available.

\subsection{Lossless Text Compression}

In the {Lossless Text Compression} task, we assess our method on the Enwik9~\cite{Singh2020WikipediaCA} and the first gigabyte of BookCorpus~\cite{Zhu2015AligningBA} datasets. In addition to GPT-2 models, we tested our method on LLAMA2-7B~\cite{touvron2023llama}. GPT-AC is benchmarked against both traditional methods, such as GZIP~\cite{rfc1952}, and contemporary neural network-based techniques like DZIP~\cite{DZIP} and TRACE~\cite{trace}. 
In the actual implementation, GPT-AC processes chunks of 2,500 characters for GPT-2 and 10,000 characters for LLAMA2-7B independently. Although this would slightly compromise the compression ratio, it enables parallel computing. 

\vspace{-0.3cm}
\begin{table}[ht!]
  \centering
  \caption{\textbf{Compression Ratio by Compression Method.} Note the compression ratio equals to \textit{Original text length} / \textit{Compressed text length}. }
  \label{tab:compression}
   \resizebox{\textwidth}{!}{
  \begin{tabular}{llcccp{0.5cm}p{0.4cm}p{0.5cm} p{0.85cm} p{0.75cm}}
    \toprule
      Model $\rightarrow$ & \multicolumn{4}{c}{GPT-AC (Ours)} & GZIP  & 7z & DZIP  & TRACE & NNCP \\
          \cmidrule(lr){2-5}
        \cmidrule(lr){3-5}
     Dataset $\downarrow$ & Llama2-7B  & GPT2-L & GPT2-M &GPT2-S  &  &   &   \\
    \midrule \midrule
    Enwik9  &\textbf{15.56} & 8.05 & 7.71 & 6.53 & 3.09 & 4.35 & 4.47 & 5.29  & 9.33  \\
    BookCorpus   &\textbf{10.55} & {8.34} & 7.89 & 7.22 & 2.77  & 3.80 & 3.95  & 4.58   & - \\
    \bottomrule
  \end{tabular}}
\end{table}

As shown in \Cref{tab:compression},  GPT-AC significantly outperforms conventional methods like GZIP and 7z in compression ratio.
Even with the GPT-2 small model, GPT-AC achieves a more than 2-fold enhancement in compression ratio compared to the widely-used GZIP, on both the Enwik9 and Book datasets. 
On Enwik9, GPT-AC with Llama2-7B records a compression ratio of 15.56, a 67\% enhancement over the previous state of the art of 9.33, based on NNCP \cite{nncp} \footnote{Refer to:\url{http://mattmahoney.net/dc/text.html}}.
As the language model increases, the compression ratio consistently improves, suggesting that larger and better-trained language models will further amplify these results. 


Note that NNCP involves updating the parameters of a large transformer model (v3.1, 199M) during encoding. 
We can also achieve an even higher compression ratio with encode-time optimization. 
However, the encode-time optimization process only enables an increase in compression ratio as the input message length goes up, and would overfit to the specific message. With random initialization, the compression ratio will be around 1.0 for the beginning of the input message, offering no benefits to similarity measurement. 

\subsection{Semantic Textual Similarity}

For the {Text Semantic Modeling}, we test the models on 
the Semantic Textual Similarity benchmark (STS-b)~\cite{cer-etal-2017-semeval}. The dataset consist of sentence pairs with labels from 0 to 5 indicating the semantic relatedness. We compare GPT-AC against GPT-2-emb~\cite{radford2019language}, where we take the last token embedding vector, and BERT-emb~\cite{devlin2019bert}, where we take the averaged token embedding vector, which has been proven to be effective in previous studies\cite{ISBERT}. We then calculate the cosine similarity between these vectors to serve as the distance measure. For GZIP, we follow~\cite{fewshotnips, jiang-etal-2023-low} and use the normalized compression distance as the metric.

\begin{table}[ht!]
  \centering
    \caption{\textbf{Semantic Textual Similarity  Performance.} Spearman Rank Correlation $\rho$ between the distance metrics and given labels for the STS datasets. $\rho * 100$ is reported.}
  \label{tab:sts}
  \begin{tabular}{llcccc}
    \toprule
    Dataset & \# Test & GPT-AC (Ours) & GZIP &  GPT-emb & BERT-emb  \\
    \midrule\midrule
    STS-12 & 3,108 & 40.2 & \textbf{50.4} & 5.4 & 30.9 \\
    STS-13 & 1,500 & \textbf{66.0}& 48.4 & 14.6 & 59.9 \\
    STS-14 & 3,750 & \textbf{55.3}& 43.3 & 10.9 & 47.7 \\
    STS-15 & 3,000 & \textbf{70.3}& 59.1 & 9.6 & 60.3 \\
    STS-16 & 1,186 &\textbf{69.5}& 59.4  & 26.8 & 60.3 \\
    STS-b & 1,379 & \textbf{55.0}& 50.7  & 12.4 & 47.3 \\
    \bottomrule
  \end{tabular}
\end{table}

As shown in \Cref{tab:sts}, our method substantially outperforms the cosine similarity distance metrics derived from GPT-2 embeddings and shows moderate enhancement over those utilizing BERT embeddings. These results demonstrate the effectiveness of the approximated information distance in capturing semantic similarities.


\subsection{Text Classification}

For {Text Classification}, we evaluate the models on PIQA (Physical Interaction: Question Answering~\cite{piqa}) and CaseHOLD (Case Holdings On Legal Decisions ~\cite{Zheng2021WhenDP}) for zero-shot classification, and SST-5(sentiment analysis)~\cite{socher-etal-2013-recursive}, Medical abstracts~\cite{medicaldataset}, AG-News(news headlines)~\cite{AGNEWS}, and Banking77(banking and financial)~\cite{Casanueva2020} for one-shot classification.
We compare our method with two main approaches: 1) fine-tuning GPT-2 or BERT with a classification layer, 2) in-context learning with GPT-2 (Refer to Appendix for detailed settings) and 3) calculate cosine similarity of Sentence-BERT (all-MiniLM-L12-v2) embeddings as a metric for classification. 

\vspace{-0.3cm}
\begin{table}[ht!]
  \centering
  \caption{\textbf{Text Classification Accuracy (100\%).} We report the averaged accuracy across 5 runs with different random seeds, together for the standard deviations. This does not apply to zero-shot experiments because the models do not contain randomness.}
  \label{tab:classification}
  \resizebox{\textwidth}{!}{
  \begin{tabular}{lllcccccc}
    \toprule
    &&Model$\rightarrow$& GPT-AC &GZIP  & GPT-prompt & GPT & BERT & (SBERT)\\
     Dataset$\downarrow$  & \# C  & Domain$\downarrow$ & (Ours) & &   &  &  &   \\
    \midrule\midrule
    \multicolumn{8}{l}{Zero-shot Multiple Choice } \\
    PIQA & 2 & Reasoning & \textbf{61.5}  & 53.4  & 50.5 & 49.2 & 50.1 & (56.5) \\
    CaseHOLD & 5 &  Legal & \textbf{58.3} & 52.4  & 20.3 & 19.9 & 35.0 & (50.6) \\
    \midrule
    \multicolumn{8}{l}{One-shot } \\
    AGNews  & 4 & News      &\textbf{47.8}$\pm$3.3      & 30.2$\pm$3.0  & 47.2$\pm$2.9 & 37.7$\pm$7.2 & 45.5$\pm$3.1 & (45.8$\pm$10.2) \\
    Medical & 5 & Bio-Med   &  \textbf{27.9}$\pm$3.2    & 25.6$\pm$2.8 & 22.1$\pm$1.2& 23.7$\pm$3.5 & 23.8$\pm$4.8 &({39.7}$\pm$9.1)\\
    SST5 & 5 & Sentiment      &  26.8$\pm$3.1  & 21.2$\pm$2.7 & \textbf{29.8}$\pm$1.6  & 22.7$\pm$2.8 & 21.1$\pm$3.3  & (26.2$\pm$2.3) \\
    Banking77 & 77 & Finance & \textbf{34.0}$\pm$1.3    & 20.3$\pm$1.5  & -  & 21.7$\pm$1.7  & 24.5$\pm$3.9 & ({53.1}$\pm$1.9)\\
    \bottomrule
  \end{tabular}}
\end{table}

As depicted in \Cref{tab:classification}, in the zero-shot multiple-choice classification context, the information distance approximated by GPT-AC delivers superior results than cosine similarity distance metrics based on the embeddings from GPT-2, BERT, and even SBERT. Note that SBERT, which is fine-tuned on 1 billion high-quality labeled sentence pairs, does not fall under our category of human-learning few-shot models; it is included to provide a point of reference.

In one-shot text classification, our method surpasses both fine-tuned GPT and BERT on all datasets.  Additionally, we also outperform the GPT-prompt version in all datasets except SST-5. Given that SST-5 is a widely used classification benchmark, we hypothesize that the superior performance of the prompt approach could be due to data leakage during GPT pre-training. Moreover, we did not apply the GPT-prompt method to the banking77 dataset because accommodating one-shot samples of 77 classes~\cite{Casanueva2020} within the GPT-2 prompt proves challenging, and adjusting the prompt can be complex. This issue represents a significant hurdle when applying GPT-2 with in-context learning.


\subsection{Text Re-ranking}

For {Text Re-Ranking}, we evaluate the models on various domain-specific zero-shot text retrieval datasets, including Trec-Covid~\cite{10.1145/3451964.3451965}, Trec-News~\cite{soboroff2019trec} , SciFact~\cite{wadden-etal-2020-fact} , BioASQ~\cite{tsatsaronis2015overview} , FiQA-2018~\cite{10.1145/3184558.3192301} , and ArguAna~\cite{wachsmuth:2018a}.
Given a query, we first retrieve the top relevant document with BM25~\cite{robertson1995okapi} with Elastic Search API\footnote{\url{https://github.com/elastic/elasticsearch}}. We then re-rank the documents with the models. We compare our system with the original BM25 ranking, and the Dense Passage Retrieval (DPR)~\cite{karpukhin-etal-2020-dense} model, a BERT-based model already fine-tuned on the MS MARCO~\cite{Campos2016MSMA} for ranking, and a text GZIP~\cite{jiang-etal-2023-low} compressor.

\vspace{-0.3cm}
\begin{table}[ht!]
  \centering
    \caption{\textbf{Text Re-Ranking Effectiveness (NDCG@10).} Retrieving top 100 relevant passages using BM25 and re-ranking using GPT-AC. We present the best result among different metric combinations. }
  \small
  \begin{tabular}{lllcccc}
    \toprule
    &&Model $\rightarrow$ & GPT-AC(Ours) & BM25  & GZIP& (DPR)\\
    \cmidrule(r){2-4}
    Dataset $\downarrow$     & \# Test & Domain &   &  &\\
    \midrule\midrule
    TREC-COVID  & 50    & COVID     & \textbf{0.694} & 0.656     & 0.447&(0.332)\\
    TREC-NEWS   & 57     & News         & 0.225 & \textbf{0.398}     & 0.142& (0.161)\\  
    SciFact     & 300   & Scientific  & 0.648& \textbf{0.665}     &0.053& (0.318)\\ 
    BioASQ      & 500    & Bio-Med       & \textbf{0.517} & 0.465     &0.157& (0.127)\\ 
    FiQA-2018   & 648    & Finance         & \textbf{0.239}& 0.236    &0.032 & (0.112)\\ 
    ArguAna     & 1406     & Argument    & \textbf{0.327} & 0.315     & 0.073& (0.175)\\ 
    \bottomrule
  \end{tabular}

  \label{tab:rerank}
\end{table}

As shown in \Cref{tab:rerank}, our proposed method outperforms the BERT-based DPR model across all settings. Despite the DPR model being fine-tuned on the massive labeled MS MARCO dataset, our performance remains superior. We also benchmark our model against the GZIP compression method. The improvements observed indicate that GPT-AC can provide significant semantic information, leading to improved ranking results. Notably, BM25 is a strong baseline in that domain-specific texts can contain many out-of-distribution terms, potentially hampering the performance of the language model. Despite this, our method demonstrates comparable performance across most of the datasets and surpasses BM25 in certain domains, particularly in Bio-Medical and Finance, which require much domain-specific understanding.


\section{Analysis}
\subsection{Comparison of Distance Metrics} 

In \Cref{tab:analysis}, we compare the results of all distance metrics with two different coding variants.
As stated in \Cref{sec:uida}, we always take the longer sequence results with $\mathcal{M}_{max}$, the shorter sequence results with $\mathcal{M}_{min}$, and the average over both of them with $\mathcal{M}_{mean}$. We aim to avoid bias towards any single sequence when both texts contain similar or equally important information, suggesting that $\mathcal{M}_{mean}$ is the most suitable. This intuition is justified by the results for STS and one-shot classification. Yet, the approach shifts for re-ranking. Typically, queries contain fewer than 20 tokens, contrasting with documents that often contain hundreds of tokens. We thus want to focus the measurement on the query part, without being disturbed by the extra or non-relevant information in the document. Therefore, $\mathcal{M}_{min}$ is the preferred choice. ArguAna, however, is a unique case as both the query and the document contain about 250 tokens with similar content, thus making $\mathcal{M}_{mean}$ more suitable. For zero-shot classification, despite the varying content and text length between the two sequences, both pieces of information remain important, making $\mathcal{M}_{mean}$ the optimal choice. Finally, we observe $\mathcal{M}_{mean}$ works better with Log-Rank, while $\mathcal{M}_{min}$ works better with Log-Prob. However, no specific trend is evident for a broader conclusion.

\begin{table}[ht!]
\caption{\textbf{Distance metric Analysis.} We also list the token length of both $x$ and $y$}
  \label{tab:analysis}
  \centering
  \resizebox{0.95\textwidth}{!}{
  \begin{tabular}{lllcccccc}
    \toprule
    &&&\multicolumn{3}{c}{Log-Prob} &\multicolumn{3}{c}{Log-Rank} \\
   \cmidrule(lr){4-6} \cmidrule(lr){7-9}
    Dataset & $len(x)$&$len(y)$&$\mathcal{M}_{max}$ & $\mathcal{M}_{min}$ & $\mathcal{M}_{mean}$  & $\mathcal{M}_{max}$ & $\mathcal{M}_{min}$ & $\mathcal{M}_{mean}$ \\
    \midrule
    \multicolumn{6}{l}{Semantic Textual Similarity (Spearman Rank Correlation $\rho * 100$)}  \\
    sts-b &13&13& 44.7& 47.6 & 48.2       & 50.4 & 54.2 & \textbf{55.0}  \\
    \midrule
     \multicolumn{6}{l}{Zero-shot Classification (Accuracy \%)} \\
    PIQA &24& 10 & 60.0 & 56.4  & \textbf{62.0}    &  59.5 & 55.0 & 61.5 \\
    CaseHold &29 &219 & 56.0 & 27.2  & 57.9         & 55.2 & 27.6 & \textbf{58.3}\\
    \midrule
    \multicolumn{6}{l}{ One-shot Classification (Accuracy \%)}\\
    AGNews &53 &53 & 41.3 & 41.1 & \textbf{47.8}   & 40.5 & 39.9 & 43.3  \\
    Medical & 296& 268 & 25.4 & 27.0 & \textbf{27.9}  & 25.5 & 26.3 & 27.7  \\
    SST5 &23&24& 24.8 & 26.2 & 26.7              & 24.3 & 26.0 & \textbf{26.8} \\
    Banking77 &15& 14& 30.6 & 25.2 & 33.9         & 29.8 & 24.7 & \textbf{34.0} \\
    \midrule
    \multicolumn{6}{l}{Re-ranking (NDCG@10)}\\
    TREC-COVID &303 &17  &0.459 &0.655 &0.467         &0.473&\textbf{0.694} &0.489 \\ 
    TREC-NEWS &808 &16   &0.173 &0.161&0.167           &0.184&0.161&\textbf{0.186}\\ %
    BioASQ  & 304 &14    &0.306&\textbf{0.517}&0.364      &0.267&0.507&0.330\\ %
    FiQA-2018 &247 &15   &0.128&\textbf{0.239}&0.154   &0.118&0.222&0.143\\ %
    ArguAna &276 &247    &0.277&0.257&0.301              &0.282&0.262&\textbf{0.307}\\ 
    SciFact &345 &21     &0.389&\textbf{0.648}&0.519     &0.351&0.635&0.478\\%
    \bottomrule
  \end{tabular}}
\end{table}


\subsection{Information Distance and Classification Accuracy} 

\begin{figure}[ht!]
\centering
\includegraphics[scale=0.5]{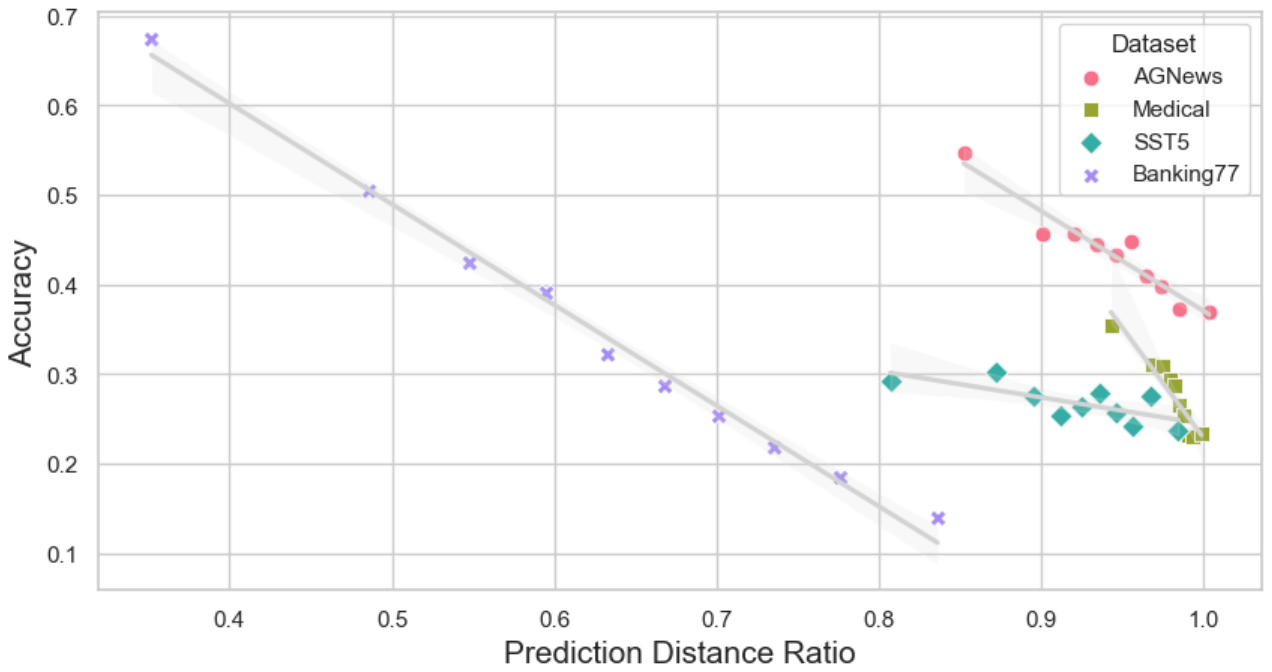} 
 \caption{\textbf{Relation between Prediction Distance Ratio and One-shot Classification Accuracy.} Experiment result under $\mathcal{M}_{mean}$ with Log-Prob.}
 \label{fig:relative}
\end{figure}

In \Cref{fig:relative}, we aim to illustrate the performance variance when the test cases have different distance scoring. For each test case, we compute the prediction distance ratio $R_{pred}(x) = \frac{\mathcal{M}(x, D_{c^*})}{\frac{1}{|C|} \sum_{c}\mathcal{M}(x, D_{c})}$. Here, $D_{c}$ represents the one-shot example in class $c$, $C$ embodies all the classes, and $c^*$ stands for the class predicted under metric $\mathcal{M}$. A smaller $R_{pred}$ suggests that the predicted class is more distant from the average distance. We then categorize all the test samples according to their $R_{pred}$ value, with each group containing 10\% of the data. In \Cref{fig:relative}, the x-axis represents the average $R_{pred}$ within each group, and the y-axis represents the group accuracy. The plot indicates that the further the predicted class deviates from the average, the better performance of our method.

\section{Conclusion and Discussion}
In this work, we introduce GPT-based Compression, a novel approach that leverages GPT models to estimate the information distance for few-shot learning. Our proposed method facilitates an out-of-the-box application of generative language models under zero-shot and one-shot scenarios without fine-tuning or prompting. This enhances the generalizability of pre-trained generative models, demonstrated by our experiments across various downstream NLP tasks. While this method can synergize with existing techniques such as further fine-tuning and various prompting techniques, we leave these combinations as potential areas for future research.

Tapping into the capabilities of pre-trained language models can bring about significant improvements in lossless text compression. Our experiments indicate a direct relationship between the scale of the language model and the improvement in compression ratio. Performance can also be elevated with fine-tuning during encoding. As pre-trained large language models become more accessible, integrating LLM-driven compression into practical applications can lead to significant advantages, including reduced storage expenses and minimized transmission overheads in various contexts.

The universal information distance is foundational, unifying various popular deep learning approaches for few-shot learning. For example, Siamese Network~\cite{koch2015siamese} uses twin networks to extract features, where $\mathcal{M}$ refers to a contrastive loss; Prototypical Network~\cite{snellprototypical} optimizes $\mathcal{M}$ to learn a better $\mathcal{D}_c$ in the embedding space; Bi-Encoder architecture used in SBERT can also be unified where $\mathcal{M}$ can be cosine similarity.

Lastly, we wish to highlight the distinction between two learning paradigms: large-data dependent learning and human-like few-shot learning. Despite the impressive ability of recent GPT models to learn from vast data, we contend that the creation of new concepts and ideas will predominantly occur in a few-shot learning manner, regardless of future advancements in these models. In the context of few-shot learning, where labeled data are scarce and cannot be used to approximate the non-computable information distance, both humans and machines are poised on an equal footing to unearth new regularities that augment compression.

\section*{Limitations} 
For downstream NLP tasks, our experiments use older versions of pre-trained language models, constrained by computational constraints and limited access. 

\newpage
\bibliographystyle{plain}
\bibliography{main}

\appendix

\newpage
\section{A Theory of Information Distance}
\label{appx:id}

The fundamental component of the information distance is the conditional Kolmogorov complexity. Formally, it's defined as follows:
\begin{definition}[\textbf{Conditional Kolmogorov Complexity}]
    $K(x|y) = \min\{|p|: U(p|y) = x\}$,
    \label{eq:cond_kol}
\end{definition} 
where $p$ is the program that converts $y$ to $x$. Put simply, \Cref{eq:cond_kol} defines the size of the shortest program with respect to Turing Machine $U$ that generates $x$ given $y$.

\begin{definition}[\textbf{Information Distance}]
    $\mathcal{E}(x, y) =\max\{K(x|y),K(y|x)\}$.
    \label{eq:id_kol}
\end{definition}


\Cref{eq:id_kol} is the formal definition of the \textit{information distance}. It describes the shortest program length that can convert from $x$ to $y$ or $y$ to $x$. To compare across different objects, we normalize the information distance defined as $\mathcal{M}_{max}(x,y)$ in Equation(4).

\begin{theorem}
    $\mathcal{E}(x,y)$ is an admissible distance metric. It is minimal in the sense that for every admissible distance $D$, we have $\mathcal{E}(x,y)\leq D(x,y) + c$ where $c$ is a constant.
    \label{thm:id}
\end{theorem}

To better understand~\Cref{thm:id}, we need two more definitions on \textit{metric} and \textit{admissible distance} according to~\cite{bennett1998information}.

\begin{definition}[\textbf{Admissible Distance}]
    A function $D:\Omega\times\Omega\rightarrow \mathbb{R}^+$ is an admissible distance if for every pair of objects $x,y\in\Omega$, the distance $D(x,y)$ is computable, symmetric and satisfies the density condition $\sum_y 2^{-D(x,y)}\leq 1$.
\end{definition}

\begin{definition}[\textbf{Metric}]
    \label{def:metric}
A distance function $D: \Omega\times\Omega \rightarrow \mathbb{R}^+$ is a \textit{metric} if it satisfies the following three criteria for any $x,y,z\in \Omega$, where $\Omega$ is a non-empty set, $\mathbb{R}^+$ represents the set of non-negative real number:

\begin{enumerate}
    \item Identity: $D(x,y) = 0$ iff $x=y$

    \item Symmetry: $D(x,y) = D(y,x)$
    
    \item Triangle Inequality: $D(x,y) \leq D(x,z)+D(z,y)$
\end{enumerate}
\end{definition}

We now prove \Cref{thm:id} based on~\Cref{lemma:uni} following ~\cite{bennett1998information}.
\begin{lemma}
\label{lemma:uni}
For every upper-semicomputable function $f(x,y)$, satisfying $\sum_{y}2^{-f(x,y)}\leq 1$, we have $K(y|x) < f(x,y)$.
\end{lemma}

\begin{proof}
    To prove that $\mathcal{E}(x,y)$ is a metric, we show it satisfies metric (in)equalities. We can infer the non-negativity and symmetry directly from the definition $\mathcal{E}(x,y)=\max\{K(x|y), K(y|x)\}$. For triangle inequality, given $x,y,z$, without loss of generality, let $\mathcal{E}(x,z)=K(z|x)$. By the self-limiting property, we have
\begin{equation}
\begin{split}
    \mathcal{E}(x,z) & = K(z|x) < K(y,z|x) < K(y|x)+K(z|x,y) \\
    & < K(y|x)+K(z|y) \leq \mathcal{E}(x,y) + \mathcal{E}(y,z).
\end{split}
\end{equation}
To prove $\mathcal{E}(x,y)$ is \textit{admissible}, we show it satisfies density requirement:
\begin{equation}
\sum\limits_{y:y\neq x}2^{-\mathcal{E}(x,y)}\leq \sum\limits_{y:y\neq x}2^{-K(y|x)}\leq 1.
\end{equation}
The second inequality is due to Kraft's inequality for prefix codes.\\
To prove the minimality, as for every admissible distance metric $D(x,y)$, it satisfies $\sum\limits_{y:y\neq x}2^{D(x,y)}\leq 1$. According to \Cref{lemma:uni}, we have $K(y|x)<D(x,y)$ and $K(x|y)<D(y,x)$.
\end{proof}





Despite of many successful applications, the $\mathcal{M}_{max}$ and $\mathcal{M}_{mean}$ introduced in Equation(4) and Equation(6) face several major practical problems. One of the major problems is the validity of triangle inequality in the condition of \textit{metric}. 
Fagin and Stockmeyer~\cite{fs} give an example of partial pattern matching where the triangle inequality does not hold. 
Veltkamp~\cite{veltkamp} puts it vividly that under the partial matching, the distance between a man and a horse is larger than the sum of the distances between a man and a centaur and between a centaur and a horse, respectively. 
The QA problems often depend on partial information and are in a similar situation as partial pattern matching.



Li~\cite{ml2006} defines the cost of conversion between $x$ and $y$ with respect to a universal Turing machine U as:
 \begin{definition}
     $\mathcal{E}_{min}(x, y) = min\{|p| : U(x, p, r) = y, U(y, p, q) = x,
 |p| + |q| + |r| \leq \mathcal{E}(x, y)\}$
 \end{definition}

Furthermore, the following theorem is proven which establishes a justifiable alternative information distance measure.
\begin{theorem}
    $\mathcal{E}_{min}(x,y) = \min \{K(x|y), K(y|x)\}$.
\end{theorem}

While $\mathcal{E}_{min}(x, y)$ is symmetric and positive, it does not satisfy the triangle inequality. But it is also universal as $\mathcal{E}(x,y)$ is universal.
Similar to normalizing $\mathcal{E}(x,y)$, we also normalize $\mathcal{E}_{min}(x,y)$, which give us $\mathcal{M}_{min}(x,y)$ as shown in Equation(5).




\section{A Theory of Human-Like Few-shot Learning}
\label{appx:hlfsl}

We follow the definition in~\cite{humanlikefewshot} and consider the following formalization of few-shot learning: 

\begin{definition}[\textbf{Few-Shot Learning}]
    Consider a universe $\Omega$, partitioned into H disjoint concept classes: $\mathcal{C}_h, h = 1, 2, \dots, H$. Few-shot ($k$-shot including zero-shot) learning can be described as follows: 
\begin{enumerate}
\item Given $n$ unlabelled samples $y_1, \cdots, y_n$ in or outside of $\Omega$;
\item Given $k$-labeled examples for each class $C_h$, for small $k$ ($k=0$ when the setting is for zero-shot);
\item The algorithm learns $\mathcal{C}_h, h = 1, 2, \cdots, H$ using the metric $\mathcal{M}$ with $n$ unlabeled samples and $k$ labeled samples, minimizing the objective function: 
\begin{equation}
    \sum^{H}_{h=1} \sum^{|\mathcal{C}_h|}_{i=1} \mathcal{M}(x_i , core_h) | y_1, \cdots, y_n, x_i \in \mathcal{C}_h, 
\end{equation}
where $core_h = \phi(k \text{ samples of } \mathcal{C}_h)$ representing a transformed representation of the k labeled samples from $\mathcal{C}_h$. 
\end{enumerate}
\end{definition}

For zero-shot learning, $core_h$ is typically another sample (i.e. to determine if they are from the same class) or description of the query. For one-shot learning, it can be the single labeled dataset. 

In order to incorporate the information of unlabeled data, we can extend the definition by adding distribution learned from any data sources available including $n$ unlabeled data and even external datasets, leading to: 
\begin{equation}
    \sum^{H}_{h=1} \sum^{|\mathcal{C}_h|}_{i=1} \mathcal{M}(x_i , x_h | \mathcal{H} (y_1, \cdots, y_n)),
\end{equation}
where $\mathcal{H} (y_1, \cdots, y_n)$ is a pre-trained model of $y_1, \cdots , y_n$, capturing the distribution.

We derive the optimal $\mathcal{M}$ from von-Neuman-Landauer Principle.

\begin{definition}[\textbf{von-Neuman-Landauer Principle}]
    Irreversibly processing 1 bit of information costs 1kT; reversible computation is free.
\end{definition}

Then for two objects $x$, $y$, the minimum energy needed to convert between x and y is: 
\begin{equation}
    \mathcal{E}_U(x, y) = min\{|p| : U(x, p) = y, U(y, p) = x\}, 
    \label{eq:vnl}
\end{equation}
where U is a universal Turing machine, assuming Church-Turing thesis. To interpret, $\mathcal{E}(x, y)$ is the length of the shortest program that reversibly converts between $x$ and $y$. These bits used in the shortest program $p$ when they are erased will cost $|p|kT$ of energy, according to the John von Neuman and Rolf
Landuaer’s law.
It's also proved in~\cite{bennett1998information} that:
\begin{equation}
    \mathcal{E}_U(x,y) = \mathcal{E}(x,y) + O(1).
\end{equation}
As we've proved in~\Cref{appx:id} that $\mathcal{E}(x,y)$ is the optimal information distance, we can rely on the adoption of $\mathcal{E}(x,y)$ to, for example, $\mathcal{M}_{max}$ to guarantee the optimality of finding $\mathcal{M}$.







\section{Adaptive Arithmetic Coding using GPT}

\begin{algorithm}[H]
    \begin{algorithmic}[1]
        \State \textbf{Input:} $T = \{t_0, t_1, t_2, ..., t_n\}$ where $t_0$ is the EOS token, GPT model $\phi$. 
        \State Initialize range $I_{low} = 0, I_{high} = 1$
        \For{ $t_i \in T, i = 1, \cdots, (n+1)$}
            \State Obtain $F_i(t_i)$ and $P_i(t_i)$ based on $\phi(t_{1:(i-1)})$
            \State $I_{low-previous} \leftarrow I_{low}$
            \State $I_{high-previous} \leftarrow I_{high}$
            \State $I_{low} \leftarrow I_{low-previous}+(I_{high-previous}-I_{low-previous})*F_i(t_i)$
            \State $I_{high} \leftarrow I_{low-previous}+(I_{high-previous}-I_{low-previous})*(F_i(t_i)+P_i(t_i))$
        \EndFor
        \State Output a single number $I_{low}\leq r< I_{high}$ as the encoded text. 
    \end{algorithmic}
    \caption{GPT-AC Encoding}
\end{algorithm}

In practice, since we cannot have float number with infinite precision. We always output the left several bits and shift the current interval by same bits once we reach the float number overflow limit.

\begin{algorithm}[H]
    \begin{algorithmic}[1]
        \State \textbf{Input:} encoded message $r$, GPT model $\phi$.  
        \State Initialize decoded text $D = \{t_0\}$ where $t_0$ is an EOS token. $i=0$
        \While{$i = 0$ or $t_i$ is not EOS token}
            \State Calculate $P_{i+1} = \phi(D)$
            \State Find index $j$ such that $F_{i+1}(v_j) \leq r< F_{i+1}(v_j) + P_{i+1}(v_j)$ where $v_i$ represents the $i$-th token in GPT vocabulary
            \State Update $t_{i+1} \leftarrow v_j$
            \State Update $D = D + \{t_{i+1}\}$
            \State Update $r \leftarrow \frac{r - F_{i+1}(v_j)}{P_{i+1}(v_j)}$
            \State Update $i \leftarrow i+1$
        \EndWhile
        \State Output D as the decoded message
    \end{algorithmic}
    \caption{GPT-AC Decoding}
\end{algorithm}


\section{Experiment Details}

We conduct our experiment on two 16G Tesla V100 GPUs, with framework and implementation adopt from HuggingFace Transformers with PyTorch.
Each experiment costs from 20min to 3h GPU time depends on the dataset size. We finish all the experiment with no more than 300 GPU hours(about 6 days).

For fine-tuning GPT and BERT with classification layers, we search within the following hyper-parameters:
\begin{itemize}[itemsep=0pt,topsep=0pt,parsep=0pt,partopsep=0pt, leftmargin=12pt]
    \item Learning Rate: 1e-5, we also tried 1e-4, 5e-5, 5e-6, 1e-6, 1e-5 gives the best learning performance
    \item Batch Size: since we are finetune with one-shot, the we select the batch size within 2, 4, 8, and restricted by the class number.
    \item Epochs: search between 500, 1000, 1500, 2000, and later the learning curve converges.
    \item Max Sequence Length: GPT2: 1024, BERT: 512
\end{itemize}

We also list all the models card we used:

\begin{itemize}[itemsep=0pt,topsep=0pt,parsep=0pt,partopsep=0pt, leftmargin=12pt]
    \item gpt-2 {https://huggingface.co/gpt2}, gpt2-medium {https://huggingface.co/gpt2-medium}, gpt2-large {https://huggingface.co/gpt2-large}, gpt2-xl {https://huggingface.co/gpt2-xl}, 
    \item BERT {https://huggingface.co/bert-base-uncased}
    \item Sentence BERT: {https://huggingface.co/sentence-transformers/all-MiniLM-L12-v2}
    
\end{itemize}


For all the dataset, we take the pre-processed datasets from hugging face, except for the Medical dataset. For the Medical dataset, we follow the official repo: https://github.com/sebischair/Medical-Abstracts-TC-Corpus.






\section{GPT In-Context Learning}\label{app:prompt}

\begin{table}[ht!]
\centering
\caption{Zero-Shot Prompts}\label{tab:casehold-prompt}
\begin{tabular}{lp{10cm}}
\toprule
 CaseHold Prompt  & Citing: \{citing\}. 
 
 Holding 0: \{holding\_0\}. 
 
 Holding 1: \{holding\_1\}. 
 
 Holding 2: \{holding\_2\}. 
 
 Holding 3: \{holding\_3\}. 
 
 Holding 4: \{holding\_4\}. 
 
 Which holding is correct (0, 1, 2, 3, or 4)?
 
 Answer: \\
\midrule

 PIQA Prompt  & Goal: \{goal\}. 
 
 Option 0: \{option\_0\}. 
 
 Option 1: \{option\_1\}. 
 
 Which option is correct (0 or 1)?
 
 Answer: \\

\bottomrule
\end{tabular}
\end{table}

\begin{table}[ht!]
\centering
\caption{One-Shot Prompt}\label{tab:agnews-prompt}
\begin{tabular}{lp{10cm}}
\toprule
 AG News Prompt  & Please classify text input into one of the following categories: World, Sports, Business, and Science/Technology. 
 
 Here are some examples:
 
    Input: \{Example 1 Text\}, Label: \{Example 1 Label\}

    Input: \{Example 2 Text\}, Label: \{Example 2 Label\}
    
 Input: \{Testing Text\}, Label:  \\
 Label Token Mapping  & 'World': 10603, 'Sports': 18153, 'Business': 24749, 'Science/Technology': 26959\\ 
\midrule

 Medical Prompt  & Please classify text input into one of the following categories: World, Sports, Business, and Science/Technology. 
 
 Here are some examples:
 
    Input: \{Example 1 Text\}, Label: \{Example 1 Label\}

    Input: \{Example 2 Text\}, Label: \{Example 2 Label\}
    
 Input: \{Testing Text\}, Label:  \\
 Label Token Mapping  & 'Neoplasms': 10603, 'Digestive system diseases': 18153, 'Nervous system diseases': 24749, 'Cardiovascular diseases': 26959,  'Pathological conditions': 0 \\ 
\midrule


 SST5 Prompt  & Please classify text input into one of the following categories: World, Sports, Business, and Science/Technology. 
 
 Here are some examples:
 
    Input: \{Example 1 Text\}, Label: \{Example 1 Label\}

    Input: \{Example 2 Text\}, Label: \{Example 2 Label\}
    
 Input: \{Testing Text\}, Label:  \\
 Label Token Mapping  & 'very negative': 10603, 'negative': 18153, 'neutral': 24749, 'positive': 26959, 'very positive': 0\\ 
\bottomrule
\end{tabular}
\end{table}

\end{document}